\documentclass[runningheads]{llncs}

\usepackage{accv}

\usepackage{accvabbrv}
\usepackage{array} 
\usepackage{graphicx}
\usepackage{booktabs}

\usepackage[accsupp]{axessibility}  % Improves PDF readability for those with disabilities.
\usepackage{cite}

\usepackage{hyperref}

\usepackage{orcidlink}

\begin{document}

\title{A Feature Generator for Few-Shot Learning} 


% Include the authors' ORCID for the camera-ready version, if at all possible.
\author{Heethanjan Kanagalingam\inst{1}\orcidlink{0009-0005-0869-5235} \and
Thenukan Pathmanathan\inst{1}\orcidlink{0009-0008-2495-7903} \and
Navaneethan Ketheeswaran\inst{1}\orcidlink{0009-0008-7622-1631} \and
Mokeeshan Vathanakumar\inst{1}\orcidlink{0009-0003-4415-8336} \and
Mohamed Afham\inst{2}\orcidlink{0000-0002-5767-9566} \and
Ranga Rodrigo\inst{1}\orcidlink{0000-0002-1034-7513}}

\authorrunning{Heethanjan et al.}

\institute{Dept. of Electronic and Telecommunication Engineering, University of Moratuwa, Sri Lanka. \and
Technical University of Darmstadt, Germany.\\
\email{heethanjanheetha@gmail.com}}

\maketitle
\begin{abstract}
 Few-shot learning (FSL) aims to enable models to recognize novel objects or classes with limited labeled data. Feature generators, which synthesize new data points to augment limited datasets, have emerged as a promising solution to this challenge. This paper investigates the effectiveness of feature generators in enhancing the embedding process for FSL tasks. To address the issue of inaccurate embeddings due to the scarcity of images per class, we introduce a feature generator that creates visual features from class-level textual descriptions. By training the generator with a combination of classifier loss, discriminator loss, and distance loss between the generated features and true class embeddings, we ensure the generation of accurate same-class features and enhance the overall feature representation. Our results show a significant improvement in accuracy over baseline methods, with our approach outperforming the baseline model by 10\% in 1-shot and around 5\% in 5-shot approaches. Additionally, both visual-only and visual + textual generators have also been tested in this paper.  The code is publicly available at \href{https://github.com/heethanjan/Feature-Generator-for-FSL}{https://github.com/heethanjan/Feature-Generator-for-FSL}.
  \keywords{Few-shot learning (FSL) \and Feature generator \and Embedding process \and Class-level semantic features }
\end{abstract}

\section{Introduction}
\begin{figure}[tb]
    \centering
    \includegraphics[height=7cm]{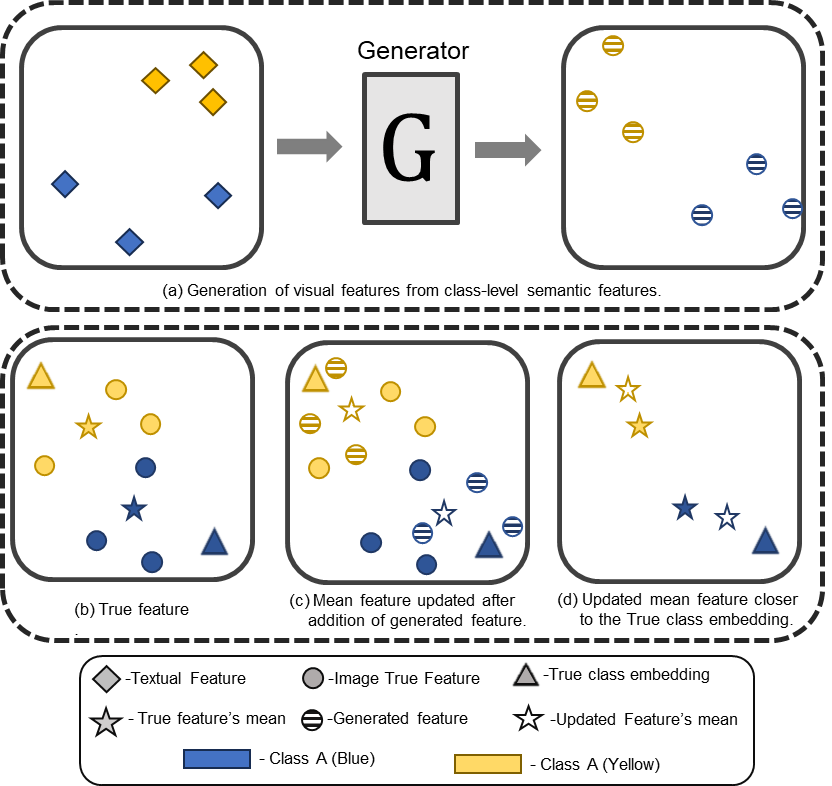}
    \caption{ The feature generation process. To generate the best optimum visual features from the class-level semantic features (a), the generated features are added to the initial true features(b), and the mean feature is updated by considering all the features (c). Finally, 
    the updated features mean, obtained by combining the generated and real features, converges closer to the true class embedding (d).} \label{fig:1}
\end{figure}

FSL is a challenging task in machine learning, where the goal is to recognize and classify objects with limited labeled data. Unlike traditional deep learning models that require large amounts of labeled data to achieve high performance, FSL aims to perform well even when only a few examples per class are available for training. This task is crucial for applications where data collection is expensive, time-consuming, or impractical, such as medical image processing, rare species identification, and personalized user experiences. 

To address the limitations of traditional models in FSL scenarios, researchers have explored various techniques, including meta-learning \cite{finn2017model, snell2017prototypical}, which aims to learn how to learn by leveraging knowledge from a wide range of tasks to adapt quickly to new ones; metric learning \cite{ vinyals2016matching}, which focuses on learning a similarity measure to effectively compare and distinguish between classes; and generative models \cite{zhang2018metagan,xian2018feature}, which can synthesize new examples to augment the limited available data.

Even though there are various models to address the issues with FSL, there remains a significant gap in effectively integrating and leveraging the complementary information from textual and visual modalities. Current methods often treat these features independently, missing the opportunity to enhance the discriminative power of support class embeddings through their combined use. Additionally, existing generative models primarily focus on augmenting visual data without fully exploiting semantic information derived from class descriptions, which could provide valuable context and improve feature generation quality.

In this study, we propose a novel approach for FSL using a feature generator that leverages semantic features to generate visual features, thereby enhancing the support class embeddings. We contribute to synthesizing visual features from class-level textual descriptions using a novel feature generator. This approach combines semantic features with visual data and uses a combined loss function to align generated features with true class embeddings closely. By generating visual features from class-level semantic features, we aim to bridge the gap between textual and visual modalities, utilizing their complementary information to denote discriminative visual features better. This ensures that the classes in the support set are correctly represented in the embedding space. \cref{fig:1} shows our approach in detail.

Here, we selected textual descriptions that accurately represent the dominant visual features of each class. These descriptions highlight the critical features that distinguish one class from another. We ensured that consistent textual descriptions were used across both the training and test sets to maintain uniformity. Additionally, whether these descriptions were manually written or automatically generated, we followed a structured format to ensure coherence throughout the dataset.

The key idea behind our approach is to generate synthetic visual features, effectively transforming the $n$-shot learning scenario into a $2n$-shot learning scenario. This allows us to fine-tune the embedding of the support set classes and capture more discriminative and accurate information. Our approach utilizes a conditional generator, which takes the semantic feature of the support class as input and generates synthesized features. These generated features are then added to the support set, enhancing their representation.

To implement our approach, we employ a feature generator architecture comprising a classifier, discriminator, and generator. The classifier is trained to classify the true features, while the discriminator is trained to differentiate between true and generated features of the image. The generator, on the other hand, takes class-level semantic features as input and generates visual features that are closely aligned with the original features. This is achieved by training the generator to minimize a combined loss function that includes classifier loss, discriminator loss, and cosine distance loss during each feature generation step. Our ablation study (Section 4.3) shows that this combined loss leads to significantly higher accuracy than other loss combinations, confirming its effectiveness. By incorporating this cumulative loss approach, our method ensures that the generated features closely align with the original ones.

To evaluate the effectiveness of our approach, we conducted experiments on the miniImageNet \cite{ravi2017optimization} and tieredImageNet \cite{ren2018meta}, which are commonly used benchmarks for FSL. We trained our generator with the Meta-Baseline \cite{chen2021meta} and Free Lunch \cite{yang2021free} baselines, and accuracy significantly improved compared to those without the generator. 

In summary, our proposed approach combines textual and visual information in the FSL setting by generating visual features from class-level semantic features.This approach not only leverages semantic features in a new way but also achieves a significant performance boost, particularly in 1-shot scenarios, which surpasses many existing state-of-the-art methods.Our experiments validate the effectiveness of our approach and demonstrate that it can be just used as a module in any baselines to improve their performance and accuracy in the few shot settings. 

 % This integration enhances the embedding process, leading to improved generalization and recognition performance. 

\section{Related Work}
\label{sec:formatting}

FSL has gained significant attention as a promising approach for classification tasks when limited labeled training data is available \cite{le2017geodesic,borowicz2019aerial,le2017co,le2021physics,le2019shadow}. FSL methods aim to leverage semantic information and generate representative features to enhance the performance of few-shot classifiers. FSL approaches can be categorized into optimization-based methods \cite{finn2017model,li2017meta,khadka2022meta,liu2020prototype} and data augmentation based methods \cite{antoniou2017data,schwartz2018encoder,wang2018low,xu2021variational}. 

Early works in Few-Shot Learning (FSL) primarily centered around optimization-based methods, which aimed to adapt models to new tasks with just a few gradient updates. These approaches, often built within meta-learning frameworks, focused on learning a good initialization for model parameters \cite{finn2017model, ravi2017optimization}.

In contrast, data augmentation-based methods sought to enhance model performance in data-scarce scenarios by generating additional training samples. For instance, FeLMi \cite{roy2022felmi} emphasizes hard mixup augmentation by interpolating between data points, while Label Hallucination \cite{jian2022label} generates labels for unseen classes. Cap2Aug \cite{roy2022cap2aug} uses captions to guide augmentation, and Global Local-Aware Augmentation \cite{shi2023global} focuses on maintaining semantic orthogonality.

Semantic features play a crucial role in bridging the gap between limited visual data and the rich information embedded in textual descriptions. Various works have explored the integration of semantic information to enhance few-shot classifiers. For instance, \cite{xing2019adapt} proposed an adaptive cross-modal FSL approach that effectively combines visual and semantic information for classification tasks. Similarly, \cite{snell2017prototypical} introduced prototype networks that utilize semantic features to generate representative prototypes for each class. Furthermore, \cite{tokmakov2019learning} explored the learning of compositional representations for few-shot recognition, emphasizing the importance of semantic information in enhancing classification performance. However, these methods primarily enhance existing visual features using semantic information rather than generating new visual features directly from textual descriptions.

Considering the implementation of generative models, \cite{zhang2018metagan} presented a novel approach for generating representative samples for few-shot classification using a conditional generative adversarial network. Their method analyzes the integration of fake samples in the FSL problem, which fails to generate more effective features, that can be solved by the semantic feature usage in the visual feature generation. 

In \cite{xu2022generating}, proposed generating representative samples for few-shot classification by leveraging semantic features and the variation autoencoder (VAE) model, demonstrating significant improvements in classification performance. Even though their method shows significant performance, their training strategy depends on the representative sample, this can become less effective when there are few or no representative samples for classes.

Considering the limitation of the method, our approach involves developing a unique generative model that generates visual features from class-level textual descriptions by considering the true feature's mean. Additionally, our method integrates a combined loss function, which is used to closely align the generated feature with the true class-level embedding.

\section{Approach}

\begin{figure*}[tb] % Use the figure* environment to span both columns
  \centering
  \includegraphics[height=6.5cm]{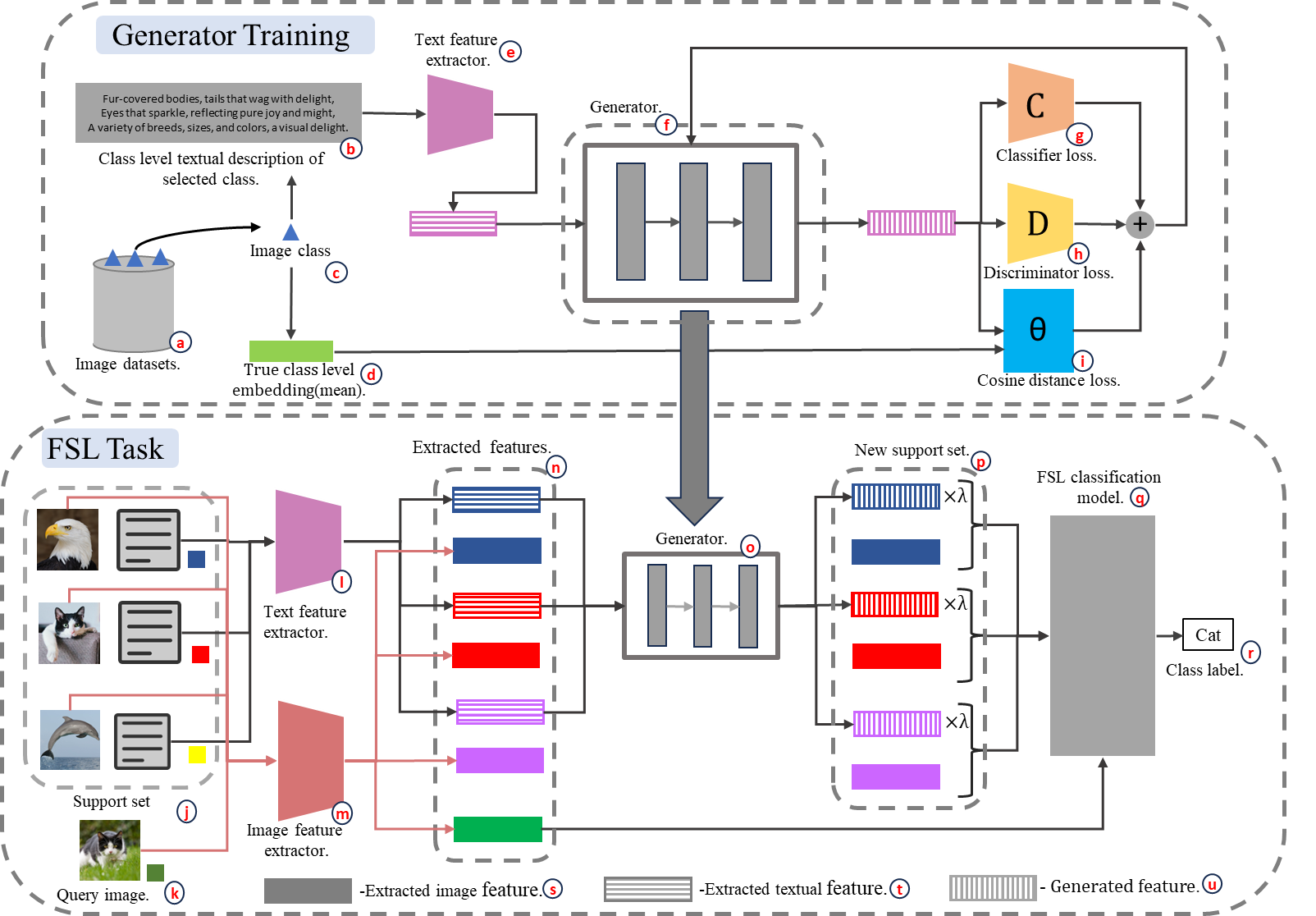} 
  \caption{The overall architecture diagram. To train the generator, an image true feature (c) and its corresponding class-level textual description (b) are taken from the image dataset (a). The true class embedding (d) is calculated by taking the mean of all the image true features belonging to the selected true feature (c) class. The generator (e) generates visual features (f) from the semantic feature (g) extracted from the class description (h) using a text feature extractor (i). The classifier loss (j)is calculated using categorical cross-entropy loss and discriminator loss (k) is calculated using binary cross-entropy loss. In contrast, the cosine distance loss (l) is computed as the distance between the true class embedding (m) and the generated feature. During training, the generator aims to minimize the sum of these three losses (j, k, l). The inference support set (n) contains Images and corresponding class descriptions. Semantic features (g) and visual features (o) are extracted using a text feature extractor (p) and an image feature extractor (q), respectively. The synthetic visual features (f) are generated by inputting the semantic features (g) to the generator (r). The generated feature is multiplied by $\lambda$ and added to the new support set (s), which is subsequently used for the FSL classification task.}
  \label{fig: overall model}
\end{figure*}

\subsection{Notation}
Suppose $N$ labeled features belonging to $n$ classes are provided for training, we can define the training dataset pairs ( $D_{tr}$), semantic training pairs ($S_{tr}$) and  the true class embedding training pairs ($T_{tr}$) are as follow:

 $D_{tr}=\{(x_0,y_0),..,(x_i,y_i),...,(x_N,y_N)\}$, where  $x_i \in X$ denotes the feature and $y_i \in Y$ is the corresponding class label. 
 
$S_{tr}=\{(s_0,l_0),...,(s_k,l_k),...,(s_n,l_n)\}$, where $s_k \in S$ denotes the class-level semantic feature and $l_k \in L$ denotes the corresponding class label. 

$T_{tr}=\{(t_0,l_0),...,(t_k,l_k),...(t_n,l_n)\}$, where
$t_k \in T$ denotes the true class embedding. 

Considering the feature generator, $G(s_k)=\widetilde{x_k}$, where $\widetilde{x_k} $ is a generated feature from the generator.

\subsection{Overall Pipeline (Textual Feature Generator)}
 \cref{fig: overall model} illustrates our generator training approach in detail. In our textual feature generator for FSL, we introduce the integration of semantic features extracted from class-level descriptions.  This integration aims to leverage the complementary information from both modalities to enhance the embedding process, as image-true features and generated features are used to get the final support feature embedding. To achieve this task, We employ a feature generator architecture, that takes a semantic feature as input and generates visual features from it. Here, $\lambda$ is used to balance the combination of the generator's output with the actual features in the support set.
\subsubsection{Class-level descriptions:} For each class, we construct a class-level description consisting of $K$ sentences that describe the classes (some examples of the descriptions are provided in the supplementary document). We employ a text encoder to convert the class-level descriptions into semantic features. The resulting $K$ semantic features from $K$ descriptions are then averaged to obtain a single class-level semantic feature, passed through the generator network to generate a visual feature. To train the generator, we adopt a multi-loss training strategy that incorporates the objectives of the classifier, discriminator, and cosine distance between the generated feature and the true class embedding of a class. Here, the true class embedding is the mean of all the image true features belonging to a particular class in the training data set. 

\subsubsection{Classifier:} The classifier is trained in parallel by inputting image true features and their corresponding labels using the categorical cross-entropy (CCE) loss between predicted class labels of the input image true features and the true class labels (\cref{eq:cross_ent}). During training, by minimizing the loss, the classifier improves its ability to classify input feature classes correctly.
The $(CCE)$ is :

\begin{equation}
CCE(P,T) = -\frac{1}{N}\sum_{i=1}^{N}\sum_{j=1}^{C} T_{ij} \log(P_{ij})
\label{eq:cat_cross_ent}
\end{equation}
Where $C$ is the number of classes, $T_{ij}$ is the target for class $j$ of sample $i$ (one-hot encoded), $P_{ij}$ is the predicted probability for class $j$ of sample $i$.

Then the classifier loss $L_c$:

\begin{equation}
L_c = CCE(C(X),Y) 
\label{eq:class_loss}
\end{equation}
where $C(x)$ is output of the classifier for input feature $x$

\subsubsection{Discriminator:} The discriminator is trained to classify between the image's true features and generated features by inputting generated features and the image's true features in parallel. To achieve this, the discriminator loss comprises two key components: the real loss and the fake loss. The real loss evaluates the discriminator's ability to classify image true features correctly. It employs the binary corss entrophy(BCE) loss, measuring the disparity between the discriminator's prediction for input image true features and the target label, which is set to 1 to represent an image's true features. Conversely, the fake loss assesses the discriminator's aptitude in distinguishing generated samples as fake. It also employs BCE loss, comparing the discriminator's prediction for inputted generated features with the target label set to 0 to denote a fake sample. The overall discriminator loss is obtained by summing the real and fake losses together. By minimizing this loss during training, the discriminator enhances its ability to differentiate between real and fake samples, contributing to the overall effectiveness of the generative model.
\\The discriminator loss ($L_d$) is:
\begin{equation}
L_d= BCE(D(X),1)+BCE(D(\widetilde{X}),0)
\label{eq:loss_bce}
\end{equation}
Where $D(x)$ is output from the discriminator for input $x$.
Here BCE is the binary cross entropy, calculated using the following equation.
\begin{equation}
{BCE(P,T)}=\frac{1}{N}\sum_{i=1}^NT_ilog(P_i)+(1-T_i)log(1-P_i)
\label{eq:cross_ent}
\end{equation}
where $P_i$ is prediction and $T_i$ is true label

\subsubsection{Cosine distance between the generated feature and the true class embedding:} Here the distance is calculated to identify how far away the generated features are from the true class embedding.
\\
cosine distance loss $(CDL)$ :
\begin{equation}
{CDL(A,B)} = \frac{1}{N}\sum_{i=1}^N\left[1-\frac{\mathbf{A}_i.\mathbf{B}_i}{\max(\lVert \mathbf{A}_i \rVert_2.\lVert \mathbf{ B}_i \rVert_2,\epsilon)}\right]
\label{eq:loss_cdl}
\end{equation}
Where $A, B$ are points in embedding space.

\subsubsection{Generator:} The overall loss function of the generator combines three important components. Firstly, the classifier loss, which uses the categorical cross entropy(CCE) loss between the predicted class labels for the input generated features and the true class labels. By incorporating this loss, the generator learns to generate features that align with the correct class. Secondly, the discriminator loss uses BCE loss between the prediction for the generated feature and a target label that is always set to 1. It guides the generator to generate features that are indistinguishable from image true features (making generated features that get classified as image true features by the discriminator). Lastly, the distance loss is computed as the cosine distance between the generated feature and the true class embedding. This loss motivates the generator to generate features that are closer to the true class embedding, allowing it to generate prominent visual features.

By incorporating semantic features and the classifier, discriminator, and embedding distance losses, our generator is trained to generate features that align with class labels and exhibit discriminability and closeness to the true embedding features. This approach enhances the embedding process in FSL, enabling improved generalization and recognition performance when presented with novel classes and limited labeled data.

Distance loss $L_\theta$:
\begin{equation}
{L_\theta=CDL(A,B)} 
\label{eq:loss_theta}
\end{equation}

\begin{equation}
{L_{\theta g}=CDL(\widetilde{x},\widetilde{T})} 
\label{eq:loss_theta_g}
\end{equation}

\begin{equation}
{L_{d g}=BCE(D(\widetilde{x}),1)} 
\label{eq:loss_dg}
\end{equation}

\begin{equation}
{L_{c g}=CCE(C(\widetilde{x}),y)}
\label{eq:loss_cg}
\end{equation}

Here, $L_{\theta g}, L_{d g} \space$ and $L_{c g}$ represent the distance loss,  discriminator loss, and classifier loss of the generated features respectively.Then the total Generator loss $L$ is given by:
\begin{equation}
L=L_{\theta g} +L_{d g}+L_{c g}
\label{eq:loss_total}
\end{equation}

\begin{equation}
L=CDL(\widetilde{x},\widetilde{T})+BCE(D(\widetilde{x}),1)+CCE(C(\widetilde{x}),y)
\label{eq:loss_final}
\end{equation}

\section{Experiments}
\subsection{Experimental Settings}

\subsubsection{Datasets:} We evaluate our method using two commonly used benchmark datasets for FSL: miniImageNet and tieredImageNet. miniImageNet is a subset of the ILSVRC-12 dataset, which is widely used for image classification tasks. It consists of 100 different classes, with each class containing 600 images. The 100 classes are split into three sets: 64 base classes for pre-training, 16 validation classes for model evaluation during training, and 20 novel classes for final testing. tieredImageNet, on the other hand, is a larger subset of the ILSVRC-12 dataset. It contains 608 classes that are sampled from a hierarchical category structure. Each class in tieredImageNet has an average of 1281 images. The dataset is first divided into 34 super-categories, which are then further split into 20 classes for training, 6 classes for validation, and 8 classes for testing. This results in a total of 351 actual categories used for training, 97 categories for validation, and 160 categories for testing. By evaluating our method on these datasets, we can assess its performance in FSL scenarios and compare it to other state-of-the-art approaches. The goal is to train models that can effectively classify images from novel classes with only a limited number of examples, mimicking the challenges of real-world FSL applications.

\subsubsection{Implementation:} 
Our generator is trained solely on the training sets of miniImageNet and tieredImageNet to avoid exposure to test set classes, ensuring consistent result comparison. The generator is initialized with weights from a visual-only generator, trained on 38,400 miniImageNet or 449,631 tieredImageNet image features, which improves its stability.

Our textual feature generator was trained separately with the train image features, which are extracted using respective feature extractors from respective baselines. For Meta-Baseline \cite{chen2021meta} and Free Lunch \cite{yang2021free}, the ResNet12 backbone is used. The dimension of the feature representation is 512 for both Meta-Baseline and Free Lunch. 
Here we used three sentences for each class, and we used the Clip\cite{radford2021learning} encoder to extract semantic features from those descriptions, and the textual feature dimension is 512. The architecture of the generator consists of three fully connected layers, each followed by a batch normalization layer and a LeakyReLU activation function.

The architecture of the discriminator consists of three fully connected layers, each followed by a batch normalization layer and a LeakyReLU activation function. The final layer projects the data from 128 dimensions to a single output, which represents the probability of the input being a real image.  The sigmoid activation function is then used to squash the output into a range between 0 and 1, representing the probability of the input being real or fake.

The architecture of the classifier consists of two fully connected layers, each followed by a batch normalization layer and a LeakyReLU activation function. The final linear layer transforms the features from the 256-dimensional space to match the number of classes in the training set, which changes for each dataset, which is 64 for miniImageNet and 351 for tiredImageNet. This transformation enables the model to output class probabilities for each input sample. The final layer uses a sigmoid activation function, which squashes the values between 0 and 1. This activation function is applied to each class probability, representing the confidence or likelihood of the input sample belonging to each class.

Here Adam optimizer is used for the generator, discriminator, and classifier models, where the initial learning rate is 1e-4 for all three. All three models are trained using  A6000 GPU for 5000 epochs at the same time as the generator is trained with the total generator loss, the discriminator is trained using the discriminator loss, and the classifier is trained using the classifier loss. It took around 10 hours to train all three models for the miniImageNet dataset.

%-----------------------------------------------------------------

\subsubsection{Baseline methods:}
Our feature generator is a simple plugin module that can be trained externally and then can be used by connecting directly with the baseline architecture. Here we used 2 baseline architectures: Meta-Baseline and Free-Lunch. In this ResNet-12 is used as the backbone. 

\subsection{Results}
\begin{table*}[ht]
\centering % This centers the table on the page
\caption{Comparison to prior works on miniImageNet and tieredImageNet.}
\label{tab:results}
\resizebox{\linewidth}{!}{%
\begin{tabular}{|c|c|cc|cc|}
\toprule
\textbf{Method} & \textbf{Backbone} & \multicolumn{2}{c|}{\textbf{miniImageNet}} & \multicolumn{2}{c|}{\textbf{tieredImageNet}} \\
 &  & \textbf{1-shot} & \textbf{5-shot} & \textbf{1-shot} & \textbf{5-shot} \\
\midrule

Matching Net \cite{NIPS2016_90e13578}&  ResNet-12& 65.64 ± 0.20& 78.72 ± 0.15 &68.50 ± 0.92 &80.60 ± 0.71 \\
MAML \cite{finn2017model} & ResNet-18 & 64.06 ± 0.18 &  80.58 ± 0.12 & - &-\\
SimpleShot \cite{wang2019revisiting} & ResNet-18 & 62.85 ± 0.20 & 80.02 ± 0.14 & 69.09 ± 0.22 & 84.58 ± 0.16 \\
CAN \cite{hou2019cross} & ResNet-12 & 63.85 ± 0.48 & 79.44 ± 0.34 & 69.89 ± 0.51 & 84.23 ± 0.37\\
S2M2 \cite{mangla2020charting} & ResNet-18 & 64.06 ± 0.18 & 80.58 ± 0.12 & - & -\\
TADAM \cite{oreshkin2018tadam} & ResNet-12 & 58.50 ± 0.30 & 76.70 ± 0.30 & 62.13 ± 0.31 & 81.92 ± 0.30\\
AM3 \cite{xing2019adapt} & ResNet-12 & 65.30 ± 0.49 & 78.10 ± 0.36 & 69.08 ± 0.47 & 82.58 ± 0.31\\
DSN \cite{simon2020adaptive} & ResNet-12 & 62.64 ± 0.66 & 78.83 ± 0.45 & 66.22 ± 0.75 & 82.79 ± 0.48\\
Variational FSL \cite{zhang2019variational} & ResNet-12 & 61.23 ± 0.26 & 77.69 ± 0.17 & - & - \\
MetaOptNet \cite{lee2019meta} & ResNet-12 & 62.64 ± 0.61 & 78.63 ± 0.46 & 65.99 ± 0.72 & 81.56 ± 0.53 \\
Robust20-distill \cite{dvornik2019diversity} & ResNet-18 & 63.06 ± 0.61 & 80.63 ± 0.42 & 65.43 ± 0.21 & 70.44 ± 0.32\\
FEAT \cite{ye2020few} & ResNet-12 & 66.78 ± 0.20 & 82.05 ± 0.14 & 70.80 ± 0.23 & 84.79 ± 0.16\\
RFS \cite{tian2020rethinking} & ResNet-12 & 62.02 ± 0.63 & 79.64 ± 0.44 & 69.74 ± 0.72 & 84.41 ± 0.55 \\
Neg-Cosine \cite{liu2020negative} & ResNet-12 & 63.85 ± 0.81 & 81.57 ± 0.56 & - &-\\
FRN \cite{wertheimer2021few} & ResNet-12 & 66.45 ± 0.19 & 82.83 ± 0.13 & 71.16 ± 0.22 & 86.01 ± 0.15\\
FeLMi \cite{roy2022felmi} & ResNet-12 & 67.47 ± 0.78 & 86.08 ± 0.44 & 71.63 ± 0.89 & 87.07 ± 0.55 \\
Label Hallucination \cite{jian2022label}& ResNet-12 & 68.28 ± 0.77 &86.54 ± 0.46& 73.34 ± 1.25& 87.68 ± 0.83 \\
Global-and Local-Aware Augmentation \cite{shi2023global} & ResNet-12 &67.25 ± 0.36 & 82.80 ± 0.30 & 72.25 ± 0.40 & 86.37 ± 0.27\\
\hline
Meta-Baseline \cite{chen2021meta} & ResNet-12 & 63.17 $\pm$ 0.23 & 79.26 $\pm$ 0.17 & 68.62 $\pm$ 0.27 & 83.74 $\pm$ 0.18\\
Free Lunch \cite{yang2021free} & ResNet-12 & 58.28 & 77.26 & 67.88 & 83.91 \\

\hline 
%Classifier-Baseline with Generator (ours) & ResNet-12 & 70.67 ± 0.21 & 79.39 ± 0.16 & \textbf{72.86 ± 0.25} & \textbf{84.11 ± 0.18}\\
Meta-Baseline with Generator (ours) & ResNet-12 & \textbf{74.62 ± 0.21} & \textbf{80.92 ± 0.16} &75.28 ± 0.25 & \textbf{89.72 ± 0.18}\\
Free Lunch with Generator (ours) & ResNet-12 &66.51 & 78.45 &\textbf{76.59 }& 84.72\\

\bottomrule
\end{tabular}%
}

\end{table*}

Following the standard setting, we conduct experiments on miniImageNet and tieredImageNet, the results and the comparison between past methods are shown in \cref{tab:results}. As most of the methods use ResNet-12 as the backbone and some with ResNet-18 with the same input image size, we can make a fair comparison between the models. 

We can see a significant accuracy improvement in the Meta baseline and Free Launch for both datasets by integrating our module. For the 1-shot we received 8.1\% to  12\% accuracy improvement and for the 5-shot we received 1.2\% to 5\% accuracy improvement. 

Here we can clearly note that the 1-shot has a significant accuracy improvement compared to the 5-shot. This is because the impact of the generated feature is high when the number of features in a support set class is less. Furthermore, we can clearly denote that, except for the 5-shot approach with the miniImageNet dataset, our method surpasses all the state-of-the-art methods by a significant margin.  

\subsection{Ablation and Analysis}
\subsubsection{Visual Generator:} Here, rather than giving the class-level textual feature as the input, we tested by giving a visual feature as the input. For this purpose, we trained the generator by adding the real extracted feature to a noise vector, which has a mean of 0.1 and a variance of 0.28. In this experiment, only the input feature is changed and we tested it in the Meta-Baseline, and we got an accuracy improvement as shown in \cref{tab:min_img_results} and \cref{tab:tiered_img_esults}.

% Your content here
\begin{table}[ht]
\caption{Comparison to prior works on miniImageNet with the integration of visual generators. Average 5-way accuracy (\%) with 95\% confidence interval.}
\label{tab:min_img_results}
\centering
\resizebox{\linewidth}{!}{%
\begin{tabular}{@{}lcc@{}}
\toprule
Model & 1-shot & 5-shot \\
\midrule

Meta-Baseline \cite{chen2021meta} & 63.17 $\pm$ 0.23 & 79.26 $\pm$ 0.17 \\

Meta-Baseline + Visual Generator (ours) & 63.64 $\pm$ 0.23 & 79.69 $\pm$ 0.16 \\

\bottomrule
\end{tabular}%
}
\end{table}

% Your content here
\begin{table}[ht]
\caption{Comparison to prior works on tieredImageNet with the integration of visual generator. Average 5-way accuracy (\%) with 95\% confidence interval.}
\label{tab:tiered_img_esults}
\centering
\resizebox{\linewidth}{!}{%
\begin{tabular}{@{}lcc@{}}
\toprule
Model & 1-shot & 5-shot \\
\midrule
Meta-Baseline \cite{chen2021meta} & 68.62 $\pm$ 0.27 & 83.74 $\pm$ 0.18 \\
Meta-Baseline + Visual Generator (ours) & 69.04 $\pm$ 0.26 & 83.43 $\pm$ 0.18 \\
\bottomrule
\end{tabular}%
}
\end{table}
%-------------------------------------------------------------------------
\subsubsection{Visual + Textual Generator:}
\begin{figure}[ht!]
    \centering
    \includegraphics[height=6.5cm]{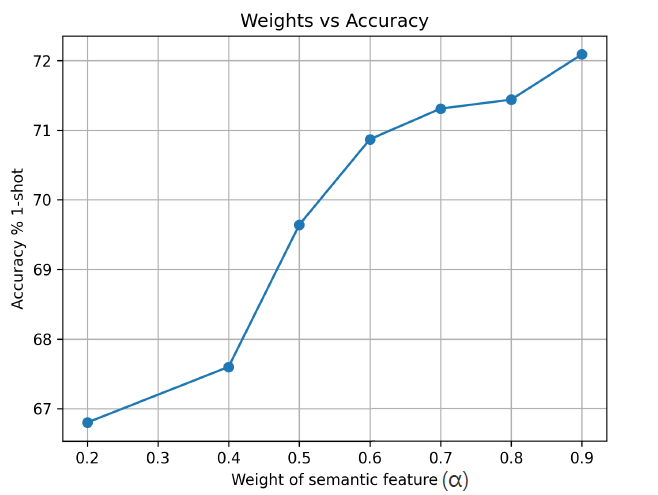}
    \caption{Classification accuracy for different semantic weight $\alpha$ in miniImageNet dataset and Meta-Baseline as the baseline }
    \label{fig:we_vs_acc}
\end{figure}

Here, rather than giving only the visual feature or the class-level textual feature, we give the features that are a combination of both visual and textual features.
% \begin{figure}[tb]
%     \centering    \includegraphics[height=6.5cm]{Images/generator.png}
%     \caption{} \label{fig:dh-protocol}
% \end{figure}
By analyzing the experimental results depicted in \cref{fig:we_vs_acc}, it can be observed that by varying the weight parameter ($\alpha$) between visual and textual features, we can manipulate the contribution of each feature type to the overall accuracy of the model. Here, $\alpha$ is used to set the weight of the textual features in the generator training in the visual + textual feature generator. When $\alpha$ increases, the accuracy increases: showing the effect of textual feature. For the textual feature generator, $\alpha$ is set to 1.

The plot illustrates a positive correlation between the weight of the textual feature and the resulting accuracy. As the weight of the textual feature increases, the model's accuracy also shows an upward trend. This suggests that the textual feature is crucial in determining the model's accuracy in this particular context.

This finding aligns with the underlying hypothesis that the textual feature contains significant information. As a result, assigning a higher weight to the textual feature enhances the strength of generated visual features. 

\subsubsection{Ablation of losses:}
We conducted the ablation for 1000 epochs and added the results in \cref{tab:loss}. The results show that all loss combinations underperform compared to our combined loss in generator training, which achieved $68.59 \pm 0.22$ at 1000 epochs, confirming its effectiveness. The original training for 5000 epochs with the combined loss reached \textbf{$74.62$} accuracy.

\begin{table}[ht]
\center
\caption{Ablation study on miniImageNet in a one-shot setting to confirm the need for our combined loss in the generator training}
\label{tab:loss}
    \begin{tabular}{|>{\centering\arraybackslash}p{0.15\linewidth}|>{\centering\arraybackslash}p{0.31\linewidth}|>{\centering\arraybackslash}p{0.3\linewidth}|>{\centering\arraybackslash}p{0.23\linewidth}|} \hline 
         \textbf{Losses} &   Classifier + Discriminator&  Cosine + Discriminator& Classifier + Cosine\\ \hline 
         \textbf{Accuracy}&   $61.07 \pm 0.22 $&  $64.19 \pm 0.22$ & $67.26 \pm 0.21$\\ \hline
    \end{tabular}
\end{table}

\subsection{Visualization of feature generator output}

% \begin{figure}[h!]
%     \centering
%     \includegraphics[height=6.5cm]{Images/plt.png}
%     \caption{Visualization of the effect of adding generated features to the support set. } \label{fig:qualt_res}
% \end{figure}

\begin{figure}[ht!]
    \centering
    \includegraphics[width=0.8\linewidth]{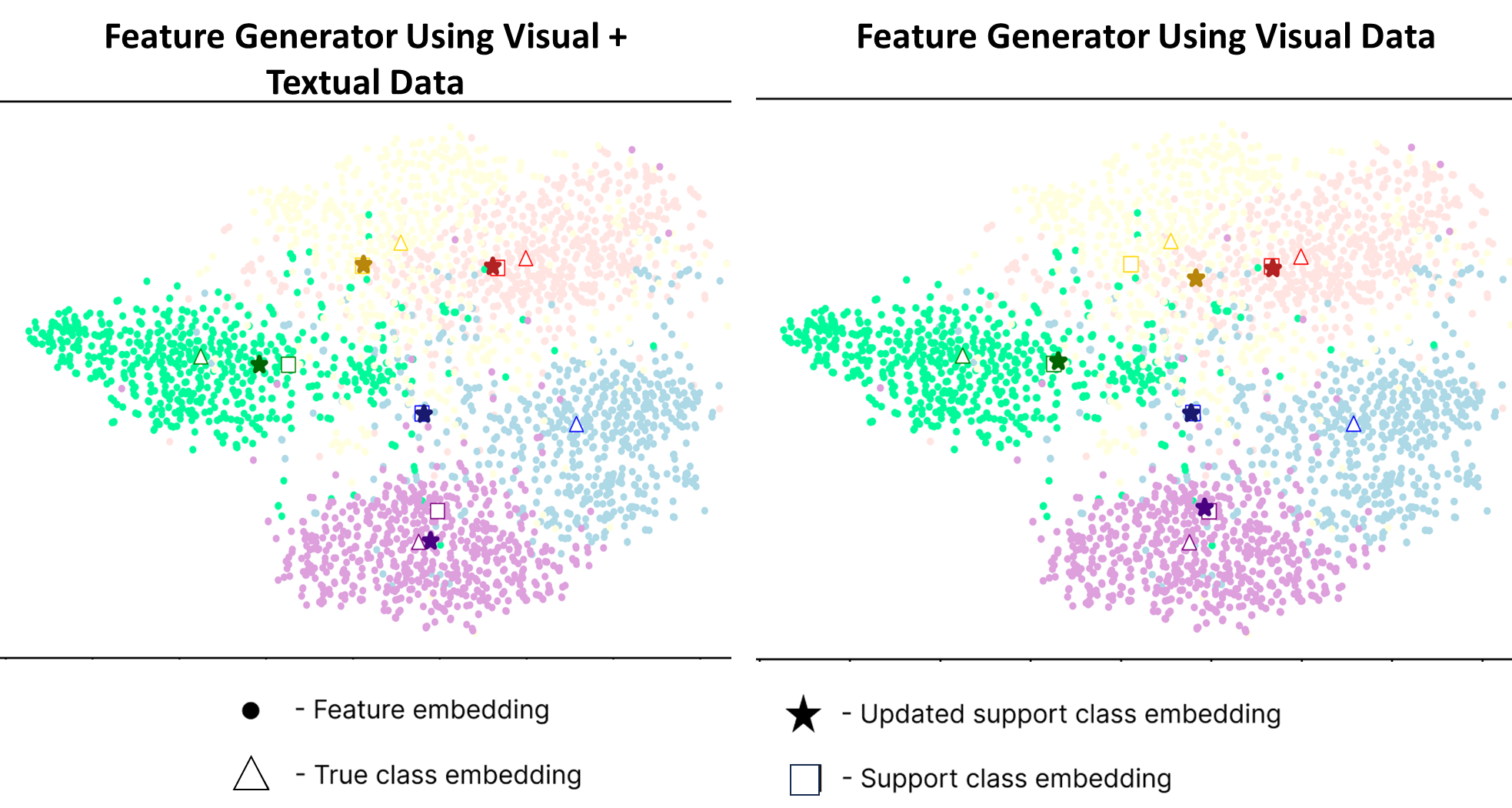}
    \caption{Visualization of the effect of using textual features for visual feature generation. Here the updated support class embeddings move closer to the true class embedding.}
    \label{fig:qualt_res}
\end{figure}
From the plots in \cref{fig:qualt_res}, we can observe that the updated support class embeddings move closer to the true class embeddings. This shift occurs when a generated feature is added to the support set, validating the significant accuracy improvement observed in the baseline models. Additionally, we can clearly observe that the features generated using visual and textual features are more effective than the only visual features, which illustrates the need for textual features.

\section{Conclusion}
In this paper, we presented an approach for FSL that integrates semantic features to enhance the embedding process for FSL tasks. We addressed the problem of inaccurate embeddings caused by few images per class by introducing a feature generator that generates visual features from textual class-level descriptions. Our approach utilized a combination of classifier loss, discriminator loss, and cosine distance loss to ensure the generation of accurate same-class features and improve the overall feature representation.

We demonstrated that the integration of semantic features significantly improves the alignment between generated and actual features, leading to better generalization and recognition performance. The integrated loss function with the generator ensures that the generated features are representative of the respective classes, thereby reducing the discrepancy between generated and true class embeddings.

Future work will focus on refining the feature generation process and exploring additional ways to incorporate semantic information. Additionally, we aim to test the scalability of our approach on more diverse and larger datasets to further confirm its robustness and general applicability.

\subsubsection{\ackname}The paper acknowledges the funds provided by the University of Moratuwa. The computational resources for this research were supported by the Accelerating Higher Education Expansion and Development (AHEAD) Operation of the Ministry of Higher Education, Sri Lanka, funded by the World Bank.

%5
\bibliographystyle{splncs04}
\bibliography{main}
\end{document}